\documentclass{article}

\usepackage{PRIMEarxiv}

\usepackage[utf8]{inputenc} 
\usepackage[T1]{fontenc}    
\usepackage{nicefrac}       
\usepackage{microtype}      
\usepackage{lipsum}
\usepackage{fancyhdr}       

\usepackage{amsmath,amssymb,amsfonts}
\usepackage{graphicx}
\usepackage{textcomp}
\usepackage{xcolor}
\usepackage{tipa}
\usepackage{rotating}
\usepackage{booktabs}

\usepackage{subcaption}
\usepackage{lineno}
\usepackage{soul}
\usepackage{multirow}
\usepackage{hyperref}

\usepackage{xurl}

\title{CIFAKE: Image Classification and Explainable Identification of AI-Generated Synthetic Images}

\author{
  Jordan J. Bird, Ahmad Lotfi \\
  Nottingham Trent University \\
  Nottingham UK\\
  \texttt{\{jordan.bird, ahmad.lotfi\}@ntu.ac.uk} \\
  }

\begin{document}
\maketitle

\begin{abstract}
Recent technological advances in synthetic data have enabled the generation of images with such high quality that human beings cannot tell the difference between real-life photographs and Artificial Intelligence (AI) generated images. Given the critical necessity of data reliability and authentication, this article proposes to enhance our ability to recognise AI-generated images through computer vision. Initially, a synthetic dataset is generated that mirrors the ten classes of the already available CIFAR-10 dataset with latent diffusion which provides a contrasting set of images for comparison to real photographs. The model is capable of generating complex visual attributes, such as photorealistic reflections in water. The two sets of data present as a binary classification problem with regard to whether the photograph is real or generated by AI. This study then proposes the use of a Convolutional Neural Network (CNN) to classify the images into two categories; Real or Fake. Following hyperparameter tuning and the training of 36 individual network topologies, the optimal approach could correctly classify the images with 92.98\% accuracy. Finally, this study implements explainable AI via Gradient Class Activation Mapping to explore which features within the images are useful for classification. Interpretation reveals interesting concepts within the image, in particular, noting that the actual entity itself does not hold useful information for classification; instead, the model focuses on small visual imperfections in the background of the images. The complete dataset engineered for this study, referred to as the CIFAKE dataset, is made publicly available to the research community for future work.
\end{abstract}


\section{Introduction}
The field of synthetic image generation by Artificial Intelligence (AI) has rapidly developed in recent years, and the ability to detect AI-generated photos has also become a critical necessity to ensure the authenticity of image data. Within recent memory, generative technology often produced images with major visual defects that were noticeable to the human eye, but now we are faced with the possibility of AI models generating high-fidelity and photorealistic images in a matter of seconds. The AI-generated images are now at the level of quality to compete with humans and win art competitions~\cite{roose2022ai}. 

Latent Diffusion Models (LDMs), a type of generative model, have emerged as a powerful tool for generating synthetic imagery~\cite{rombach2022high}. These recent developments have brought about a paradigm shift in our understanding of creativity, authenticity, and truth. This has led to a situation wherein there is consumer-level technology available that could quite easily be used for the violation of privacy and to commit fraud. These philosophical and societal implications are at the forefront of the current state of the art, raising fundamental questions about the nature of trustworthiness and reality. Recent technological advances have enabled the generation of images with such high quality that human beings cannot tell the difference between a real-life photograph and an image which is no more than a hallucination of an artificial neural network's weights and biases. 

Generative imagery that is indistinguishable from photographic data raises questions both ontological, those which concern the nature of being, and epistemological, surrounding the theories of methods, validity, and scope. Ontologically, given that humans cannot tell the difference between images from cameras and those generated by AI models such as an Artificial Neural Network, in terms of digital information, \textit{what is real and what is not?} The epistemological reality is that there are serious questions surrounding the reliability of human knowledge and the ethical implications that surround the misuse of these types of technology. The implications suggest that we are in growing need of a system that can aid us in the recognition of real images versus those generated by AI. 

This study explores the potential of using computer vision to enhance our newfound inability to recognise the difference between real photographs and those that are AI-generated. Given that there are many years worth of photographic datasets available for image classification, these provide examples for a model of real images. Following the generation of a synthetic equivalent to such data, we will then explore the outputs of the model before finally implementing methods of differentiation between the two types of images. 

There are several scientific contributions with multidisciplinary and societal implications arising from this study. Firstly, a dataset, referred to as CIFAKE, is generated with latent diffusion and released to the research community. The CIFAKE dataset provides a contrasting set of real and fake photographs and contains 120,000 images (60,000 images from the existing CIFAR-10\footnote{Collection of images that are commonly used to train machine learning and computer vision algorithms available from: \url{https://www.cs.toronto.edu/~kriz/cifar.html}} dataset and 60,000 images generated for this study), making it a valuable resource for researchers in the field. Second, this study proposes a method for enhancing our waning human ability to recognise AI-generated images through computer vision, using the CIFAKE dataset for classification. Finally, this study proposes the use of Explainable AI (XAI) to further our understanding of the complex processes involved in synthetic image recognition, as well as the visualisation of the important features within those images. These scientific contributions provide important steps forward in addressing the modern challenges posed by rapid developments of modern technology, and have important implications for ensuring the authenticity and trustworthiness of data. 

The remainder of this article is as follows; the state-of-the-art research background is initially explored in Section \ref{sec:background} with a discussion of relevant related studies in the field. Following this, the methodology followed by this study is then detailed in Section \ref{sec:method},  which provides technical implementation and the method followed for binary classification of real versus AI-generated imagery. The results of these experiments are presented with discussion in Section \ref{sec:results} before this work is finally concluded, and future work is proposed in Section \ref{sec:conclusion}.  

\section{Background}
\label{sec:background}

The ability to distinguish between real imagery and those which are generated by machine learning models is important for a number of reasons. Identification of real data provides confirmation on the authenticity and originality of the image; for example, a fine-tuned Stable Diffusion Model (SDM) could be used to generate a synthetic photograph of an individual committing a crime or vice versa, providing false evidence of an alibi for a person who was, in reality, otherwise elsewhere. Misinformation and \textit{fake news} is a significant modern problem, and machine-generated images could be used to manipulate public opinion~\cite{pennycook2021psychology,singh2022predicting}. Situations where synthetic imagery is used in fake news can promote its false credibility and have serious consequences~\cite{bonettini2021use}. Cybersecurity is another major concern, with research noting that synthetically generated human faces can be used in false acceptance attacks and have the potential to gain unauthorised access to digital systems~\cite{deb2020advfaces,khosravy2021model}. In \cite{BIRD2023170}, it was observed that synthetically generated signatures could overcome signature verification systems with ease. 

Latent Diffusion Models is a new approach for generating images which use attention mechanisms and a U-Net to reverse the process of Gaussian noise and, ultimately, use text conditioning to generate novel images from random noise. Details on the methodological implementation of LDMs can be found in Section \ref{sec:method}. The approach is rapidly developing but young, and thus literature on the subject is currently scarce. The models are a new approach in the field of generative models; thus, the literature is young, and few applications have been explored. Examples of notable models include Dall-E by OpenAI~\cite{ramesh2021zero}, Imagen from Google~\cite{saharia2022photorealistic}, and the open source equivalent, SDM from StabilityAI~\cite{rombach2022high}. These models have pushed the boundaries of image quality, both in realism, and arguably artistic ability. This has led to much debate on the professional, social, ethical, and legal considerations of technology~\cite{roose2022ai}. 

The majority of research in the field is cutting-edge and is presented as preprints and recent theses. In \cite{chambon2022adapting}, researchers proposed to train SDM on medical imaging data, achieving images of a higher quality that could potentially lead to increased model abilities via data augmentations. It is worth mentioning that in \cite{schneider2023mo,schneider2023thesis}, diffusion models were found to have the ability to generate audio as well as images. In 2021, the results of Yi et al.\cite{yi2021exploring} suggested that diffusion models were highly capable of generating realistic artworks, fooling human subjects into believing that the works were created by human beings. Given this, researchers have noted that diffusion models have a promising capability in co-creating with human artists ~\cite{guo2023artverse}. 

DE-FAKE, proposed by Sha et al.~\cite{sha2022fake}, shows that images generated by various latent diffusion approaches may contain digital fingerprints to suggest they are synthetic. While visual glitches are growingly rare given the advances of model quality, it may be possible that computer vision approaches will detect these attributes within images that the human eye cannot. The Fourier transforms presented in \cite{corvi2022detection} show visual examples of these digital fingerprints.

When discussing the topic of vision, the results in \cite{amerini2019deepfake} suggest that optical flow techniques could detect synthetic human faces within the FaceForensics dataset with around 81.61\% accuracy. Extending to the temporal domain, \cite{guera2018deepfake} proposes recurrence in AI-generated video recognition achieving 97.1\% accuracy over 80 frames due to minor visual glitches at the pixel-scale. In Wang et al.\cite{wang2022m2tr}, EfficientNets and Vision Transformers are proposed within a system that can detect forged images by adversarial models at an F1 score of 0.88 and AUC of 0.95, competing with the state of the art on the DeepFake Detection Challenge dataset while remaining efficient. In the aforementioned study, a Convolutional Neural Network was used to extract features, similarly to the approach proposed in this study, prior to processing through attention-based approaches. 

Similarly, convolutional and temporal techniques were proposed in \cite{saikia2022hybrid} to achieve 66.26\% to 91.21\% accuracy in a mixed set of synthetic data detection datasets. Chrominance components $CbCr$ within a digital image were noted in \cite{li2020identification} as a promising route for the detection of minor pixel disparities that are sometimes present within synthetic images.

Human recognition of manipulation within images is waning as a direct result of image generation methods improving. A study from Nightingale et al.\cite{nightingale2017can} in 2017 suggested that humans find difficulty in recognising photographs that have been edited by image processing techniques. Since this study, there have been nearly five years of rapid development in the field to date. 

The review of the relevant literature has highlighted the rapid developments within AI-generated imagery and the challenges posed today with respect to their detection. Generative models have enabled the generation of high-fidelity, photorealistic images within a matter of seconds that humans often cannot distinguish between when compared to reality. This conclusion sets the stage for the studies presented by this work and argues the need to fill the gap in the knowledge when it comes to the availability of examples of synthetic data. 

\section{Method}
\label{sec:method}
This section describes the methods followed by this study in terms of their technical implementation and application for the detection of synthetic images. This section first describes the collection of data for the real data and then the methodology followed to generate the synthetic equivalent for comparison. Sections \ref{subsec:real} and \ref{subsec:synth} will describe how 60,000 images are collected and 60,000 images synthetically generated, respectively. This forms the overall dataset of 120,000 images. Section \ref{subsec:image-classification} will then describe the machine learning model engineering approach which aims to recognise the authenticity of the images, before Section \ref{subsec:xai} notes the approach for Explainable AI to interpret model predictions. 

\subsection{Real Data Collection}
\label{subsec:real}

For the class label ``REAL'', interpreted as a positive class value ``1'', data is collected from the CIFAR-10 dataset~\cite{krizhevsky2009learning}. It is a dataset of $60,000$, $32 \times 32$ RGB images of real subjects divided into ten classes. The classes within the dataset are airplane, automobile, bird, cat, deer, dog, frog, horse, ship and truck. There are $6,000$ images per class. For each class 5,000 images are used for training and $1,000$ for testing, i.e. a testing dataset of 16.6\%. Within this study, all images from the training dataset are used for the training of positive class ``REAL''. Therefore, $50,000$ are used for training and $10,000$ for testing. 
\begin{figure}
    \centering
    \includegraphics[scale=1]{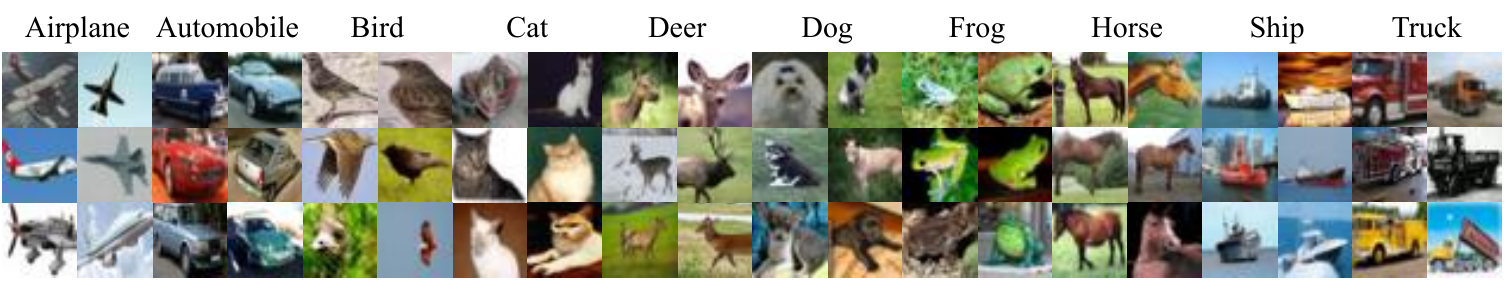}
    \caption{Examples of images from the CIFAR-10 image classification dataset~\cite{krizhevsky2009learning}.}
    \label{fig:cifar-examples}
\end{figure}
Samples of images within the CIFAR-10 dataset that form the ``REAL'' class can be observed in Figure \ref{fig:cifar-examples}.


\subsection{Synthetic Data Generation}
\label{subsec:synth}
The synthetic images generated for this study use SDM\footnote{\url{https://huggingface.co/CompVis/stable-diffusion-v1-4}}, a form of LDM that is open source for the public. 
The goal is to model the diffusion of image data through a latent space given a textual context. If noise, such as that of a Gaussian distribution, is iteratively added to an image, the image ultimately becomes noise and all prior visual information is lost. To generalise the reverse of this process is to, therefore, generate a synthetic image from noise. The method of reverse diffusion can be put simply as, given an image $x$ at timestep $t$, $x_t$, output the prediction of $x_{i-1}$ through the prediction of noise and subsequent removal by classical means. 

A noisy image $x_t$ is generated from the original $x_0$ by the following:

\begin{equation}
    \label{eq:noise}
    x_t=\sqrt{\bar{\alpha}_t} x_0+\sqrt{1-\bar{\alpha}_t} \varepsilon,
\end{equation}

\noindent where noise is $\varepsilon$, and the adjustment according to the time step $t$ is $\bar{\alpha}$. 
The method of this study is to make use of the reverse process of 50 noising steps, which from $x_{50}$ will ultimately form $x_0$, the synthetic image. The neural network $\varepsilon_\theta$ thus minimises the following loss function:

\begin{equation}
Loss = E_{t, x_0, \varepsilon} \left[ \left| \left| \varepsilon - \varepsilon_\theta(x_t, t) \right| \right| ^2 \right].
\end{equation}

Further technical details on the approach can be obtained from~\cite{rombach2022high}.The model chosen for this approach is Stable Diffusion 1.4, which is trained in the LAION2B-en, LAION-high-resolution and LAION-aesthetics v2.5 + datasets\footnote{\url{https://laion.ai/blog/laion-5b/}}. The aforementioned datasets are a cleaned subset of the original LAION-5B dataset~\cite{schuhmann2022laion}, which contains 5.85 billion text-image pairs. 

\begin{table}[] \footnotesize
\centering
\caption{Latent Diffusion prompt modifiers for generating the 10-class synthetic dataset. All prompts are preceded by ``a photograph of \{a/an\}'' and modifiers are used equally for the $6000$ images. }
\label{tab:prompts}
\begin{tabular}{@{}ll@{}}
\toprule
\textbf{Class Label}         & \textbf{Prompt Modifiers}                                                                                                              \\ \midrule
\textit{\textbf{Airplane}}   & :- aircraft, airplane, fighter, flying, jet, plane                                                                                        \\
\textit{\textbf{Automobile}} & :- family, new, sports, vintage                                                                                                           \\
\textit{\textbf{Bird}}       & \begin{tabular}[c]{@{}l@{}}:- flying, in a tree, indoors, on water, outdoors, \\ walking\end{tabular}                                     \\
\textit{\textbf{Cat}}        & \begin{tabular}[c]{@{}l@{}}:- indoors, outdoors, walking, running, eating, \\ jumping, sleeping, sitting\end{tabular}                     \\
\textit{\textbf{Deer}}       & \begin{tabular}[c]{@{}l@{}}:- herd, in a field, in the forest, outdoors, running, \\ wildlife photography\end{tabular}                    \\
\textit{\textbf{Dog}}        & \begin{tabular}[c]{@{}l@{}}:- indoors, outdoors, walking, running, eating, \\ jumping, sleeping, sitting\end{tabular}                     \\
\textit{\textbf{Frog}}       & \begin{tabular}[c]{@{}l@{}}:- European, in the forest, on a tree, on the ground, \\ swimming, tropical, wildlife photography\end{tabular} \\
\textit{\textbf{Horse}}      & \begin{tabular}[c]{@{}l@{}}:- herd, in a field, in the forest, outdoors, running, \\ wildlife photography\end{tabular}                    \\
\textit{\textbf{Ship}}       & \begin{tabular}[c]{@{}l@{}}:- at sea, boat, cargo, cruise, on the water, river, \\ sailboat, tug\end{tabular}                             \\
\textit{\textbf{Truck}}      & \begin{tabular}[c]{@{}l@{}}:- 18-wheeler, car transport, fire, garbage, heavy goods, \\ lorry, mining, tanker, tow\end{tabular}           \\ \bottomrule
\end{tabular}
\end{table}

The SDM is used to generate a synthetic equivalent to the CIFAR-10 dataset which contains $6,000$ images of ten classes. The classes are airplane, automobile, bird, cat, deer, dog, frog, horse, ship and truck. Following observations from the CIFAR-10 dataset, this study implements prompt modifiers to increase diversity of the synthetic dataset, which can be observed in Table \ref{tab:prompts}. As with the real dataset, $50,000$ images are used for the training data and $10,000$ for testing data, provided with a class label to denote that the image is not real.

\subsection{Image Classification}
\label{subsec:image-classification}
Image classification is an algorithm that predicts a class label given an input image. The learnt features are extracted from the image and processed in order to provide an output, in this case, whether or not the image is real or synthetic. This subsection describes the selected approach to classification.  

In this study, the Convolutional Neural Network (CNN)\cite{lecun2015deep,gu2018recent,li2021survey} is employed to learn from the input images. It is   
the concatenation of two main networks with intermediate operations. These are the convolutional layers and the fully connected layers. The initial convolutional network within the overall model is the CNN, which can be operationally generalised for an image of dimensions $x$ and a filter matrix $w$ as follows:
\begin{equation}
    (x * w)(i, j) = \sum_{m=1}^{M}\sum_{n=1}^{N} x(i+m-1, j+n-1)w(m,n),
\end{equation}
where $(i,j)$ is the output for the feature map, and $(m,n)$ represents the location of the filter $w$. The output is derived by applying convolutional operations to the input $x$ with each of the filters (which are learnable) and applying an activation function $f$, which, in the context of this study, is the Rectified Linear Unit (ReLu) $f(x) = \max(0, x)$. 

For an image of $(height, width)$ dimensions and filters depending on the filter kernel of $(height_{kernel})$ and $(width_{kernel})$ with a $stride = 1$ and no padding for simplicity, the output would have dimensions:

\begin{equation}
    (height - height_{kernel} + 1, width - width_{kernel} + 1).
\end{equation}

\noindent Then, a pooling operation is performed to reduce the spatial dimensions and flatten the output to then input into the densely connected layers. For $L = HWD$ (height, width and dimensions), the one-dimensional flattened output vector is simply $x = [x_1, x_2, ..., x_L]$. The output vector $y$ is ultimately output from the dense layer(s) as $y = f(W_{L}+ b)$, for the weight matrix $W$ and the bias $b$. The activation function $f$ in this study, as in CNN, is the ReLu activation function $f(x) = \max(0, x)$. 

The goal of the network in this study is to classify whether the image is a real photograph or an image generated by a LDM, and thus is a problem of binary classification. Therefore, the output of the network is a single neuron with the S-shaped Sigmoid activation function:

\begin{equation}
\sigma(x) = \frac{1}{1+e^{-x}} 
\end{equation}

The ``FAKE'' class is 0 and the ``REAL'' class is 1, therefore, a value closer to either of the two values represents the likelihood of that class. Although this aids in learning as it is differentiable, the values are rounded to the closest value for inference. 

Although the goal of the network is to use backpropagation to reduce binary cross-entropy loss, this study also notes an extended number of classification metrics. These are the Precision, which is a measure of how many of the predictive positive cases are positive, a metric which allows for the analysis of false-positives:

\begin{equation}
\text{Precision} = \frac{\text{True positives}}{\text{True positives + False positives}}.
\end{equation}

\noindent The Recall which is a measure of how many positive cases are correctly predicted, which enables analysis of false-negative predictions:

\begin{equation}
    \text{Recall} = \frac{\text{True positives}}{\text{True positives + False negatives}},
\end{equation}

\noindent This measure is particularly important in this case, as it is in fraud detection, since a false negative would falsely accuse the author of generating their image with AI. Finally, the F-1 score is considered:

\begin{equation}
    \text{F1 score} = 2 \times \frac{\text{Precision} \times \text{Recall}}{\text{Precision} + \text{Recall}},
\end{equation}

\noindent which is a unified metric of precision and recall. \\

The dataset that forms the classification is the collection of real images and the equivalent synthetic images generated, detailed in Sections \ref{subsec:real} and \ref{subsec:synth}, respectively. $100,000$ images are used for training ($50,000$ real images and $50,000$ synthetic images), and $20,000$ are used for testing ($10,000$ real and $10,000$ synthetic).

Initially, CNN architectures are benchmarked as a lone feature extractor. That is, the filters of $\{16,32,64,128\}$ are benchmarked in layers of $\{1,2,3\}$, flattened, and connected directly to the output neurone. The highest-performing feature extractor topology is then used to benchmark the highest-performing dense network featuring $\{32,64,128,256,...,4096\}$ rectified linear units in layers of $\{1,2,3\}$. These 36 artificial neural networks are then compared with regard to classification metrics to derive the topology that performs best. 

\subsection{Explainable AI}
\label{subsec:xai}
While deep learning approaches often lead to impressive predictive ability, many algorithms are black boxes that provide no reasoning for their classification. The aim of Explainable AI (XAI) is to extract meaning from algorithms and provide readable interpretations of why a prediction or decision is being made\cite{gunning2019xai}. Regarding the experiments in this work, the CNN simply predicts an image \textit{is} real or synthetic, and then XAI is used to provide interpretations as to \textit{why} the image is real or synthetic. 

Given that the literature shows that humans have a major difficulty in recognising synthetic imagery, it is important to display and visualise minor defects within the image that could suggest that it is not real.

The method selected for explainable AI (XAI) and interpretation is Gradient Class Activation Mapping (Grad-CAM)\cite{selvaraju2017grad}. Grad-CAM interprets the gradients of the predicted class along with the CNN feature maps, which can therefore be spatially localised with respect to the original input (the image) and produce a heatmap. This is generated by the Recitified Linear Unit (ReLU) function as:
\begin{equation}
    L_{Grad-CAM}^{c} = ReLU(\sum_k\alpha_kA^k), 
\end{equation}
where $\alpha_k$ is the global average pooling $\frac{1}{Z}\sum_i\sum_j\frac{\partial y_c}{\partial A_{i,j}^k}$ of spatial locations $Z$, and $\frac{\partial y_c}{\partial A_{i,j}^k}$ are the gradients of the model. 

The approach is used for interpretation in the final step of this study, given the random data selected from the two classes. Due to the nature of heatmapping, the results of the algorithm are visually interpreted with discussion.

\subsection{Experimental Hardware and Software}
\label{subsec:hardwaresoftware}
The neural networks used for the detection of AI-generated images were engineered with the TensorFlow library~\cite{tensorflow2015-whitepaper}. All TensorFlow seeds were set to 1 for replicability. The Latent Diffusion model used for the generation of synthetic data was Stable Diffusion version 1.4~\cite{rombach2022high}; Random seed vectors were denoised for a total of 50 steps to form images and the Euler Ancestral scheduler was used. Synthetic images were rendered at a resolution of 512px before resizing to 32px by bilinear interpolation to match the resolution of CIFAR-10.

All algorithms in this study were executed using a single Nvidia RTX 3080Ti GPU, which has 10,240 CUDA cores, a clock speed of 1.67 GHz, and 12GB GDDR6X VRAM.

\section{Results and Observations}
\label{sec:results}
This section presents examples of the dataset followed by the findings of the planned computer vision experiments. The dataset is also released for the public research community for use in future studies, given the important implications of detecting AI-generated imagery\footnote{The Dataset can be downloaded from: \\\url{https://www.kaggle.com/datasets/birdy654/cifake-real-and-ai-generated-synthetic-images}}.

\subsection{Dataset Exploration}

\begin{figure}
    \centering
    \includegraphics[scale=0.78]{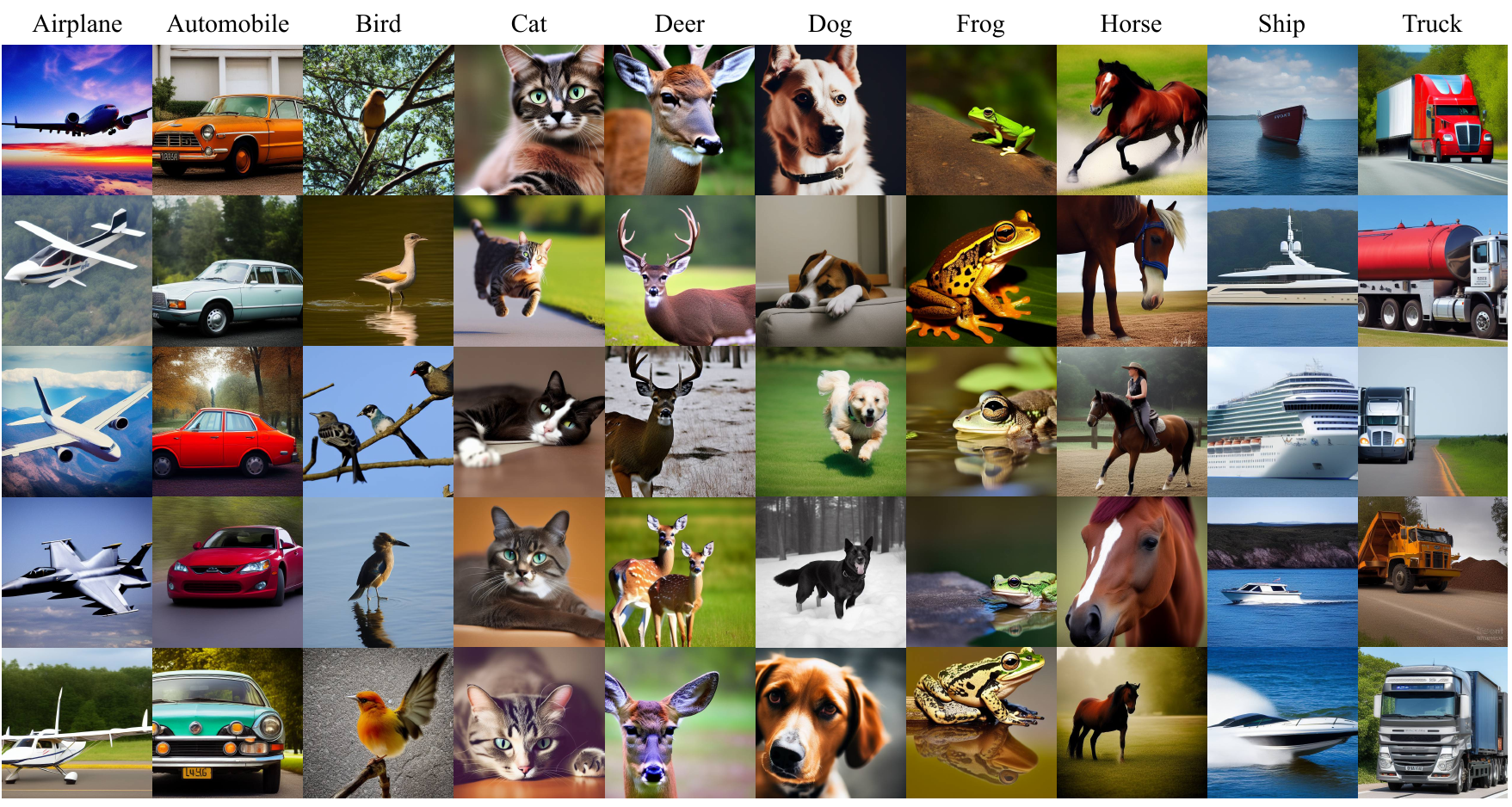}
    \caption{Examples of AI-generated images within the dataset contributed by this study, selected at random with regards to their real CIFAR-10 equivalent labels.}
    \label{fig:examples-synth}
\end{figure}

Random samples of images used in this study and within the provided dataset can be observed in Figure \ref{fig:examples-synth}. Five images are presented for each class label, and all of the images within this figure are synthetic, which have been generated by the SDM. Note within this sample that the images are high-quality and, for the most part, seem to be difficult to discern as synthetic by the human eye. Synthetic photographs are representative of their counterparts from reality and feature complex attributes such as depth of field, reflections, and motion blur.

\begin{figure}
    \centering
    \includegraphics{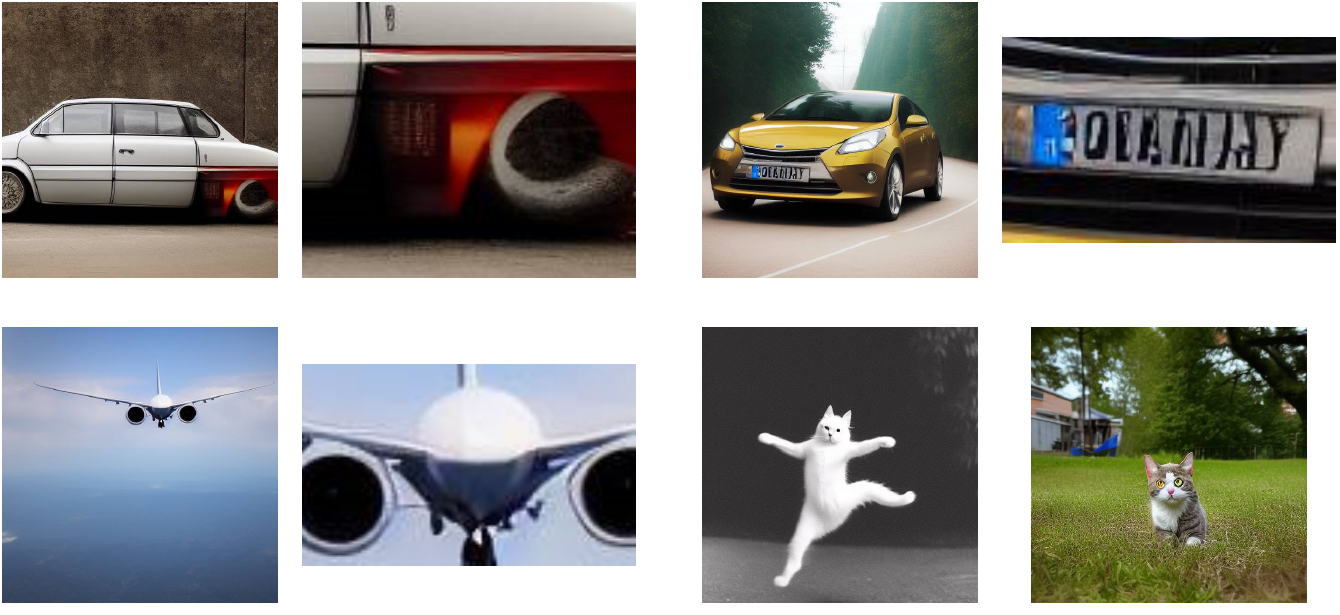}
    \caption{Examples of visual defects found within the synthetic image dataset.}
    \label{fig:defects}
\end{figure}

It can also be observed that there are visual imperfections within some of the images. Figure \ref{fig:defects} shows a number of examples of the data set where the model has output images with visual glitches. Given that the LAION dataset provides physical descriptions of the image content, little to no information on text is provided, and thus, it can be seen that the model produces shapes similar to alphabetic characters. Also observed here is a lack of important detail, such as the case of a jet aircraft that has no cockpit window. It seems that this image has been produced by combining the knowledge of jet aircraft (in particular, the engines) along with the concept of an Unmanned Aerial Vehicle's chassis. Finally, there are also some cases of anatomical errors for living creatures, seen in these examples through the cat's limbs and eyes. 

\begin{figure}
    \centering
    \includegraphics[scale=0.9]{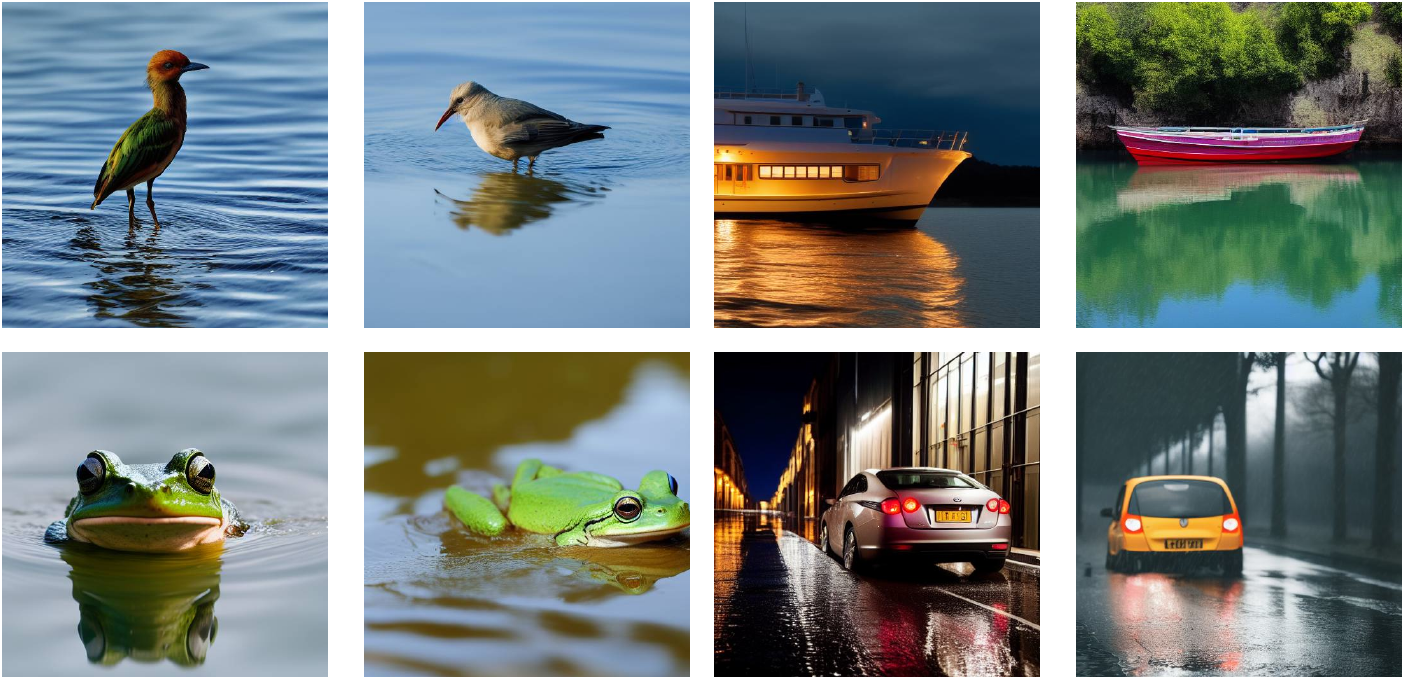}
    \caption{A selection of AI-generated images within the dataset. Examples of complex visual attributes generated by the latent diffusion model which include realistic water and reflections. }
    \label{fig:reflections}
\end{figure}

Complex visual concepts are present within much of the dataset, with examples shown in Figure \ref{fig:reflections}. Observe that the ripples in the water and reflections of the entities are highly realistic and match what would be expected within a photograph. In addition to complex lighting, there is also evidence of depth of field and photographic framing.

\subsection{Classification Results}
In this subsection, we present the results for the computer vision experiments for image classification. The problem faced by the CNN is that of binary classification, whether or not the image is a real photograph or the output of an LDM.

\begin{table}[] \footnotesize
\centering
\caption{Observed classification accuracy metrics for feature extraction networks. }
\label{tab:cnn-acc}
\begin{tabular}{@{}llll@{}}
\toprule
\multirow{2}{*}{\textbf{Filters}} & \multicolumn{3}{l}{\textbf{Layers}}                             \\ \cmidrule(l){2-4} 
                                  & \textit{\textbf{1}} & \textit{\textbf{2}} & \textit{\textbf{3}} \\ \cmidrule(r){1-1}
\textit{\textbf{16}}              & 90.06               & 91.46               & 91.63               \\
\textit{\textbf{32}}              & 90.38               & \textbf{92.93}      & 92.54               \\
\textit{\textbf{64}}              & 90.94               & 92.71               & 92.38               \\
\textit{\textbf{128}}             & 91.39               & 92.98               & 92.07               \\ \bottomrule
\end{tabular}
\end{table}

\begin{table}[] \footnotesize
\centering
\caption{Observed validation loss for the filters within the Convolutional Neural Network.}
\label{tab:cnn-loss}
\begin{tabular}{@{}llll@{}}
\toprule
\multirow{2}{*}{\textbf{Filters}} & \multicolumn{3}{l}{\textbf{Layers}}  \\ \cmidrule(l){2-4} 
                                  & \textbf{1} & \textbf{2} & \textbf{3} \\ \cmidrule(r){1-1}
\textbf{16}                       & 0.254      & 0.222      & 0.21       \\
\textbf{32}                       & 0.237      & 0.18       & 0.193      \\
\textbf{64}                       & 0.226      & 0.196      & 0.219      \\
\textbf{128}                      & 0.234      & 0.221      & 0.259     
         \\ \bottomrule
\end{tabular}
\end{table}

The results validation accuracy and loss metrics for the feature extractors can be found in Tables \ref{tab:cnn-acc} and \ref{tab:cnn-loss}, respectively. All feature extractors scored relatively highly without the need for dense layers to process the feature maps, with an average classification accuracy of 91.79\%. The lowest loss feature extractor was observed to use two layers of 32 filters, which led to an overall classification accuracy of 92.93\% and a binary cross-entropy loss of 0.18. The highest accuracy model, two layers of 128 filters, scored 92.98\% with a loss of 0.221. 

\begin{table}[] \footnotesize
\centering
\caption{Observed validation precision for the filters within the Convolutional Neural Network.}
\label{tab:cnn-precision}
\begin{tabular}{@{}llll@{}}
\toprule
\multirow{2}{*}{\textbf{Filters}} & \multicolumn{3}{l}{\textbf{Layers}}                             \\ \cmidrule(l){2-4} 
                                  & \textit{\textbf{1}} & \textit{\textbf{2}} & \textit{\textbf{3}} \\ \cmidrule(r){1-1}
\textit{\textbf{16}}              & 0.903               & 0.941               & 0.921               \\
\textit{\textbf{32}}              & 0.878               & 0.923               & 0.937               \\
\textit{\textbf{64}}              & 0.908               & 0.947               & 0.936               \\
\textit{\textbf{128}}             & 0.92                & 0.948               & 0.94       \\ \bottomrule
\end{tabular}
\end{table}

\begin{table}[] \footnotesize
\centering
\caption{Observed validation recall for the filters within the Convolutional Neural Network.}
\label{tab:cnn-recall}
\begin{tabular}{@{}llll@{}}
\toprule
\multirow{2}{*}{\textbf{Filters}} & \multicolumn{3}{l}{\textbf{Layers}}                             \\ \cmidrule(l){2-4} 
                                  & \textit{\textbf{1}} & \textit{\textbf{2}} & \textit{\textbf{3}} \\ \cmidrule(r){1-1}
\textit{\textbf{16}}              & 0.897               & 0.885               & 0.911               \\
\textit{\textbf{32}}              & 0.938               & 0.936               & 0.912               \\
\textit{\textbf{64}}              & 0.92                & 0.904               & 0.91                \\
\textit{\textbf{128}}             & 0.906               & 0.909               & 0.898               \\ \bottomrule
\end{tabular}
\end{table}

\begin{table}[] \footnotesize
\centering
\caption{Observed validation F1-Score for the filters within the Convolutional Neural Network.}
\label{tab:cnn-f1}
\begin{tabular}{@{}llll@{}}
\toprule
\multirow{2}{*}{\textbf{Filters}} & \multicolumn{3}{l}{\textbf{Layers}}                             \\ \cmidrule(l){2-4} 
                                  & \textit{\textbf{1}} & \textit{\textbf{2}} & \textit{\textbf{3}} \\ \cmidrule(r){1-1}
\textit{\textbf{16}}              & 0.9                 & 0.912               & 0.916               \\
\textit{\textbf{32}}              & 0.907               & 0.93                & 0.924               \\
\textit{\textbf{64}}              & 0.91                & 0.925               & 0.923               \\
\textit{\textbf{128}}             & 0.913               & 0.928               & 0.919               \\ \bottomrule
\end{tabular}
\end{table}

Extended validation metrics are presented in Tables \ref{tab:cnn-precision}, \ref{tab:cnn-recall}, and \ref{tab:cnn-f1} which detail the validation precision, recall, and F1-scores, respectively. The F1 score, which is a unification of precision and recall, had a mean value of 0.929 with the highest being 0.936. A small standard deviation of 0.003 was observed. 

Following these experiments, the lowest loss feature extractor is selected for further engineering of the network topology. This was the model which had two layers of 32 convolutional filters. 

\begin{table}[] \footnotesize
\centering
\caption{Observed validation accuracy for the dense layers within the Convolutional Neural Network.}
\label{tab:dense-acc}
\begin{tabular}{@{}llll@{}}
\toprule
\multirow{2}{*}{\textbf{Neurons}} & \multicolumn{3}{l}{\textbf{Layers}}                             \\ \cmidrule(l){2-4} 
                                  & \textit{\textbf{1}} & \textit{\textbf{2}} & \textit{\textbf{3}} \\ \cmidrule(r){1-1}
\textit{\textbf{32}}              & 93.2                & 92.84               & 92.96               \\
\textit{\textbf{64}}              & 93.55               & 92.73               & 93.26               \\
\textit{\textbf{128}}             & 92.99               & 93.29               & 93.18               \\
\textit{\textbf{256}}             & 92.97               & 92.88               & 92.88               \\
\textit{\textbf{512}}             & 93.05               & 92.58               & 93.33               \\
\textit{\textbf{1024}}            & 92.9                & 92.91               & 92.75               \\
\textit{\textbf{2048}}            & 92.78               & 92.76               & 92.7                \\
\textit{\textbf{4096}}            & 92.62               & 92.52               & 92.88               \\ \bottomrule
\end{tabular}
\end{table}

\begin{table}[] \footnotesize
\centering
\caption{Observed validation loss for the dense layers within the Convolutional Neural Network.}
\label{tab:dense-loss}
\begin{tabular}{@{}llll@{}}
\toprule
\multirow{2}{*}{\textbf{Neurons}} & \multicolumn{3}{l}{\textbf{Layers}}                             \\ \cmidrule(l){2-4} 
                                  & \textit{\textbf{1}} & \textit{\textbf{2}} & \textit{\textbf{3}} \\ \cmidrule(r){1-1}
\textit{\textbf{32}}              & 0.186               & 0.182               & 0.187               \\
\textit{\textbf{64}}              & 0.182               & 0.193               & 0.177               \\
\textit{\textbf{128}}             & 0.187               & 0.183               & 0.178               \\
\textit{\textbf{256}}             & 0.187               & 0.192               & 0.194               \\
\textit{\textbf{512}}             & 0.188               & 0.193               & 0.184               \\
\textit{\textbf{1024}}            & 0.199               & 0.194               & 0.192               \\
\textit{\textbf{2048}}            & 0.194               & 0.2                 & 0.219               \\
\textit{\textbf{4096}}            & 0.234               & 0.204               & 0.19                \\ \bottomrule
\end{tabular}
\end{table}

The results for the engineering of the overall network are presented in Tables \ref{tab:dense-acc} and \ref{tab:dense-loss}, which contain the validation accuracy and loss, respectively. The lowest loss observed was 0.177 binary cross-entropy when the CNN was followed by three layers of 64 rectified linear units. The highest accuracy, on the other hand, was 93.55\%, which was achieved when implementing a single layer of 64 neurons. 

\begin{table}[] \footnotesize
\centering
\caption{Observed validation precision for the dense layers within the Convolutional Neural Network.}
\label{tab:dense-precision}
\begin{tabular}{@{}llll@{}}
\toprule
\multirow{2}{*}{\textbf{Neurons}} & \multicolumn{3}{l}{\textbf{Layers}}                             \\ \cmidrule(l){2-4} 
                                  & \textit{\textbf{1}} & \textit{\textbf{2}} & \textit{\textbf{3}} \\ \cmidrule(r){1-1}
\textit{\textbf{32}}              & 0.932               & 0.916               & 0.929               \\
\textit{\textbf{64}}              & 0.925               & 0.92                & 0.93                \\
\textit{\textbf{128}}             & 0.948               & 0.942               & 0.935               \\
\textit{\textbf{256}}             & 0.939               & 0.926               & 0.931               \\
\textit{\textbf{512}}             & 0.944               & 0.924               & 0.946               \\
\textit{\textbf{1024}}            & 0.933               & 0.94                & 0.939               \\
\textit{\textbf{2048}}            & 0.942               & 0.922               & 0.929               \\
\textit{\textbf{4096}}            & 0.926               & 0.914               & 0.923               \\ \bottomrule
\end{tabular}
\end{table}

\begin{table}[] \footnotesize
\centering
\caption{Observed validation recall for the dense layers within the Convolutional Neural Network.}
\label{tab:dense-recall}
\begin{tabular}{@{}llll@{}}
\toprule
\multirow{2}{*}{\textbf{Neurons}} & \multicolumn{3}{l}{\textbf{Layers}}                             \\ \cmidrule(l){2-4} 
                                  & \textit{\textbf{1}} & \textit{\textbf{2}} & \textit{\textbf{3}} \\ \cmidrule(r){1-1}
\textit{\textbf{32}}              & 0.932               & 0.943               & 0.93                \\
\textit{\textbf{64}}              & 0.948               & 0.936               & 0.935               \\
\textit{\textbf{128}}             & 0.91                & 0.923               & 0.973               \\
\textit{\textbf{256}}             & 0.919               & 0.932               & 0.926               \\
\textit{\textbf{512}}             & 0.915               & 0.928               & 0.919               \\
\textit{\textbf{1024}}            & 0.925               & 0.916               & 0.915               \\
\textit{\textbf{2048}}            & 0.912               & 0.934               & 0.925               \\
\textit{\textbf{4096}}            & 0.926               & 0.939               & 0.936               \\ \bottomrule
\end{tabular}
\end{table}

\begin{table}[] \footnotesize
\centering
\caption{Observed validation F1-Score for the dense layers within the Convolutional Neural Network.}
\label{tab:dense-f1}
\begin{tabular}{@{}llll@{}}
\toprule
\multirow{2}{*}{\textbf{Neurons}} & \multicolumn{3}{l}{\textbf{Layers}}                             \\ \cmidrule(l){2-4} 
                                  & \textit{\textbf{1}} & \textit{\textbf{2}} & \textit{\textbf{3}} \\ \cmidrule(r){1-1}
\textit{\textbf{32}}              & 0.932               & 0.929               & 0.93                \\
\textit{\textbf{64}}              & 0.936               & 0.928               & 0.933               \\
\textit{\textbf{128}}             & 0.928               & 0.932               & 0.932               \\
\textit{\textbf{256}}             & 0.929               & 0.929               & 0.929               \\
\textit{\textbf{512}}             & 0.929               & 0.926               & 0.932               \\
\textit{\textbf{1024}}            & 0.929               & 0.928               & 0.927               \\
\textit{\textbf{2048}}            & 0.927               & 0.928               & 0.927               \\
\textit{\textbf{4096}}            & 0.926               & 0.926               & 0.929               \\ \bottomrule
\end{tabular}
\end{table}

\begin{figure}
    \centering
    \includegraphics[width=12cm]{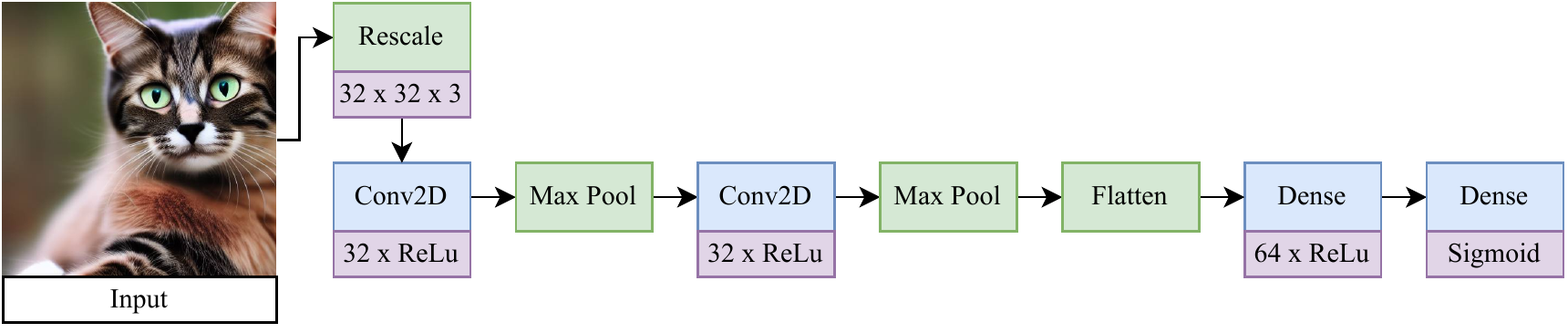}
    \caption{An example of one of the final model architectures following hyperparameter search for the classification of real or AI-generated images.}
    \label{fig:final-model}
\end{figure}

Additional validation metrics for precision, recall, and F-1 score are also provided in Tables Tables \ref{tab:dense-precision}, \ref{tab:dense-recall}, and \ref{tab:dense-f1}, respectively. Similarly to the prior experiments, the standard deviation of F1-scores was relatively low at 0.003. The highest F-1 score was the network that used a single dense layer of 64 rectified linear units, with a value of 0.936. 
The aforementioned highest F1 score model is graphically detailed in Figure \ref{fig:final-model} to provide a visual example of the network topology. 

\begin{figure}
    \centering
    \includegraphics[width=12cm]{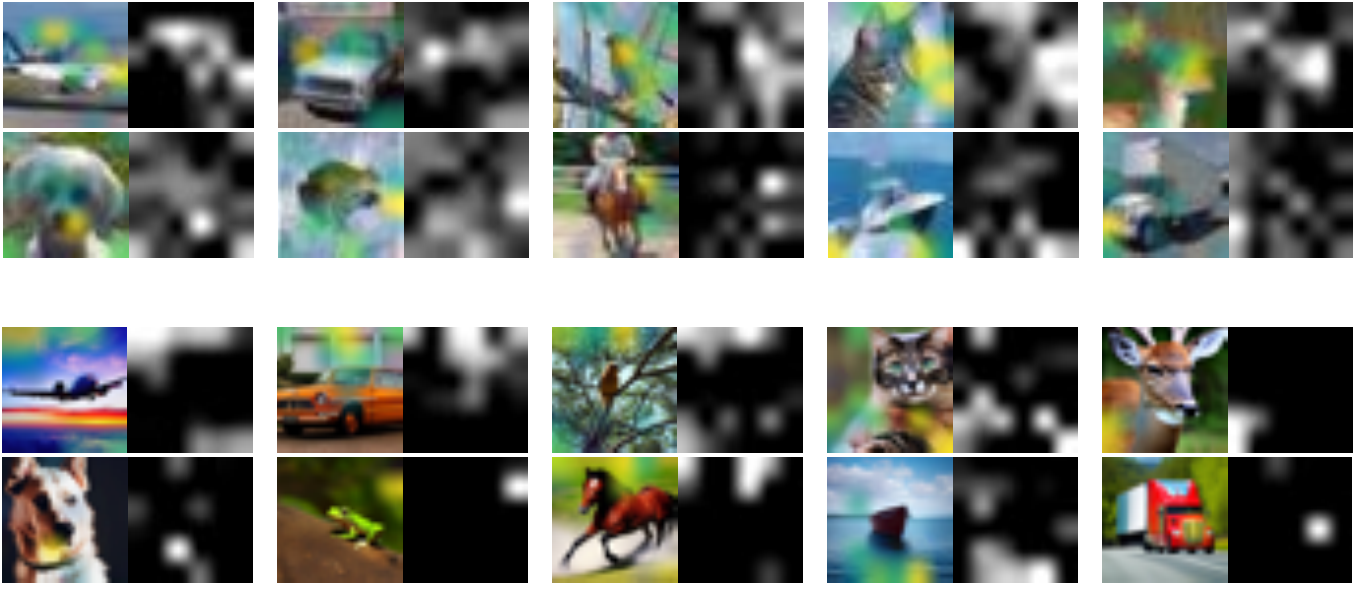}
    \caption{Gradient class activation maps (Grad-CAM) overlays and raw heatmaps for prediction interpretation. Top examples show real images and bottom examples show AI-generated images. Brighter pixels represent features contributing to the output class label.}
    \label{fig:gradcam}
\end{figure}

Figure \ref{fig:gradcam} shows examples of the interpretation of predictions via Grad-CAM. Brighter pixels in the image represent areas that contribute more to the decision of the CNN. It can be observed that there is a significantly different distribution of features given the binary classification problem. Firstly, the classification of real images can be interpreted as a more holistic approach in which the majority of the contents of the image are useful for recognition. On the other hand, the classification of synthetic images is somewhat more atomistic and sparse. Note that for the recognition of AI-generated imagery, activation occurs in select parts of the image that are more likely to present visual glitches that are difficult to recognise with the human eye. An example of this can be seen for the image of the frog, where an out-of-focus bokeh is the only attribute that suggests the image is not real. For the truck, only the pattern of the radiator grill is considered useful for classification. 

The XAI approach also shows an interesting mechanic in a more general sense. Given the examples of airplane, bird, frog, horse and ship, note that the object within the image has little to no class activation overlay whatsoever. This suggests that the actual focus of the image itself, the entity, contains almost no useful features for synthetic image recognition. This suggests that the model is often available to produce a near-perfect representation of the entity.

\section{Conclusion and Future Work}
\label{sec:conclusion}
This study has proposed a method for enhancing our waning ability to recognise AI-generated images through the use of Computer Vision and to provide insight into predictions with visual cues. To achieve this, this study proposed the generation of a synthetic dataset with Latent Diffusion, recognition with Convolutional Neural Networks, and interpretation through Gradient Class Activation Mapping. Results showed that the synthetic images were high-quality and featured complex visual attributes, and that the binary classification could be achieved with around 92.98\% accuracy. Interpretation by Grad-CAM revealed interesting concepts within the images that were useful for predictions.

In addition to the method proposed within this study, a significant contribution is made through the release of the CIFAKE dataset. The dataset contains a total of $120,000$ images ($60,000$ real images from CIFAR-10 and 60,000 synthetic images generated for this study). The CIFAKE dataset provides the research community a valuable resource for future work on the societal problems faced due to the rapid development of AI-generated imagery. The dataset provides a significant expansion of the resource availability for the development and testing of applied computer vision approaches for this problem.

The reality of AI generating images that are indistinguishable from real-life photographic images raises fundamental questions about the limits of human perception, and thus this study proposed to enhance that ability by \textit{fighting fire with fire}. The proposed approach addresses the challenges of ensuring the authenticity and trustworthiness of visual data. 

Future work could involve exploring further techniques for classification of the dataset provided. For example, the implementation of attention-based approaches are a promising new field that could provide increased ability and an alternative method of explainable AI. In addition, with even further improvements to synthetic imagery in the future, it is important to consider updating the dataset with images generated by these approaches. Additionally, consideration of generating images from other domains, such as human faces and clinical scans, would provide additional datasets for this type of study and expand the applicability of our proposed approach to other fields of research. 

To finally conclude, this study provides contributions to the ongoing implications of AI-generated images. The proposed approach supports important implications of ensuring data authenticity and trustworthiness, providing not only a system that can recognise synthetic images, but also supporting data and interpretation. The public release of the CIFAKE dataset generated within this study provides a valuable resource for interdisciplinary research.

\bibliography{bibliography}
\bibliographystyle{ieeetr}

\end{document}